\documentclass[letterpaper,10pt,conference,twoside]{ieeeconf}

\usepackage[T1]{fontenc}
\usepackage[english]{babel}
\usepackage{amssymb}
\usepackage{amsmath}
\usepackage{graphicx}
\usepackage{bm}
\usepackage{algorithm}
\usepackage{algorithmic}
\usepackage{xcolor}
\usepackage{url}
\usepackage{dsfont}

\usepackage{microtype}
\usepackage[subtle,tracking=normal]{savetrees}
\usepackage[hidelinks]{hyperref}

\newcommand{\up}[1]{^{\text{#1}}}
\newcommand{\down}[1]{_{\text{#1}}}
\newcommand{\norm}[1]{\left\lVert#1\right\rVert}
\newcommand{\R}{\mathbb{R}}
\newcommand{\N}{\mathbb{N}}

\newcommand{\argmin}{\operatornamewithlimits{argmin}}

\usepackage[noadjust]{cite}

\makeatletter
\let\old@ps@headings\ps@headings
\let\old@ps@IEEEtitlepagestyle\ps@IEEEtitlepagestyle
\def\confheader#1{%
    \def\ps@headings{%
        \old@ps@headings%
        \def\@oddhead{\strut\hfill#1\hfill\strut}%
        \def\@evenhead{\strut\hfill#1\hfill\strut}%
    }%
    \def\ps@titlepagestyle{%
        \def\@oddhead{\strut\hfill#1\hfill\strut}%
        \def\@evenhead{\strut\hfill#1\hfill\strut}%
    }%
    \ps@headings%
}
\makeatother
\confheader{%
    \parbox{20cm}{\vspace{-2cm}\textit{Accepted Preprint at 2025 IEEE-RAS 24th International Conference on Humanoid Robots (Humanoids), August 2025}}
}

\title{
Extremum Flow Matching for\\Offline Goal Conditioned Reinforcement Learning
}
\author{
Quentin Rouxel, Clemente Donoso, Fei Chen, Serena Ivaldi, and Jean-Baptiste Mouret
\thanks{%
Q. Rouxel, C. Donoso, S. Ivaldi, and J. B. Mouret are with Inria, CNRS, Universit\'e de Lorraine, France ({\tt\footnotesize firstname.lastname@inria.fr}). Q. Rouxel, and F. Chen are
with the Department of Mechanical and Automation Engineering, T-Stone
Robotics Institute, The Chinese University of Hong Kong, Hong Kong ({\tt\footnotesize quentinrouxel@cuhk.edu.hk, f.chen@ieee.org}).}%
\thanks{
This work was supported by the CPER CyberEntreprises, the Creativ’Lab platform of Inria/LORIA, the EU Horizon project euROBIN (GA n.101070596), the France 2030 program through project PI3 (ANR-22-EXOD-007, ANR-22-EXOD-004), the French National Research Agency under France 2030 via the ENACT AI Cluster (ANR-23-IACL-0004).
This work is also supported in part by the Research Grants Council of the Government of the Hong Kong Special Administrative Region (SAR) via Grant 24209021, Grant 14213324, and Grant C7100-22GF, and in part by the InnoHK of the Government of the Hong Kong SAR via the Hong Kong Centre for Logistics Robotics.}%
}

\IEEEoverridecommandlockouts
\begin{document}
\maketitle


\begin{abstract}
Imitation learning is a promising approach for enabling generalist capabilities in humanoid robots, but its scaling is fundamentally constrained by the scarcity of high-quality expert demonstrations.
This limitation can be mitigated by leveraging suboptimal, open-ended play data, often easier to collect and offering greater diversity. 
This work builds upon recent advances in generative modeling, specifically Flow Matching, an alternative to Diffusion models. 
We introduce a method for estimating the minimum or maximum of the learned distribution by leveraging the unique properties of Flow Matching, namely, deterministic transport and support for arbitrary source distributions. 
We apply this method to develop several goal-conditioned imitation and reinforcement learning algorithms based on Flow Matching, where policies are conditioned on both current and goal observations. We explore and compare different architectural configurations by combining core components, such as critic, planner, actor, or world model, in various ways. We evaluated our agents on the OGBench benchmark and analyzed how different demonstration behaviors during data collection affect performance in a 2D non-prehensile pushing task. Furthermore, we validated our approach on real hardware by deploying it on the Talos humanoid robot to perform complex manipulation tasks based on high-dimensional image observations, featuring a sequence of pick-and-place and articulated object manipulation in a realistic kitchen environment. Experimental videos and code are available at: \url{https://hucebot.github.io/extremum_flow_matching_website/}
\end{abstract}


\section{Introduction}

In recent years, imitation learning has reemerged as a powerful approach to solving complex manipulation tasks on real robots \cite{florence2022implicit, chi2023diffusion, black2024pi_0}. Unlike reinforcement learning, which typically relies on simulation and sim-to-real transfer, imitation learning bypasses the need for simulation, reward engineering, and domain adaptation by directly learning behaviors from teleoperated demonstrations (i.e., behavior cloning). This allows policies to handle environments that are difficult to simulate, such as dexterous or nonprehensile manipulation involving complex contacts, friction, and interactions with articulated, soft, or deformable objects. Moreover, end-to-end training on pixel-based inputs makes it possible to train policies for a wide range of tasks without any feature engineering. However, imitation learning comes at the cost of requiring large and diverse demonstration datasets, a process that is slow and resource intensive to collect on real hardware.

As the aim of this line of research is to build fully generalist policies capable of solving various tasks, it becomes essential for users to specify desired goals. This need has driven the rise of goal-conditioned imitation learning \cite{ding2019goal, reuss2023goal}. Goals can be provided in various forms, through discrete labels (which require costly dataset labeling and segmentation), goal images \cite{cui2022play, sridhar2024nomad}, or, more recently, natural language instructions \cite{reuss2024multimodal} via vision-language-action models \cite{black2024pi_0}.

\begin{figure}[t]
    \centering
    \includegraphics[trim=0cm 0cm 0cm 0cm,clip,width=1.0\linewidth]{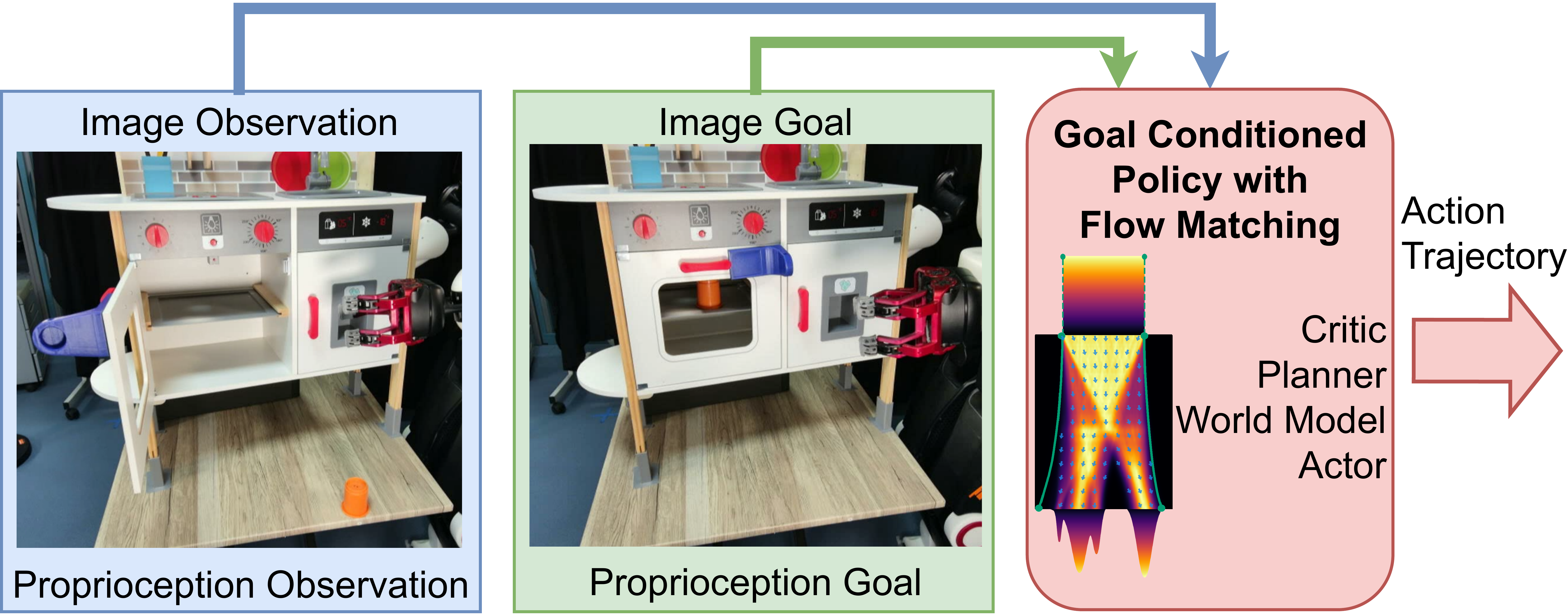}\\
    \vspace{0.2cm}
    \includegraphics[trim=0cm 0cm 0cm 0cm,clip,height=4.8cm]{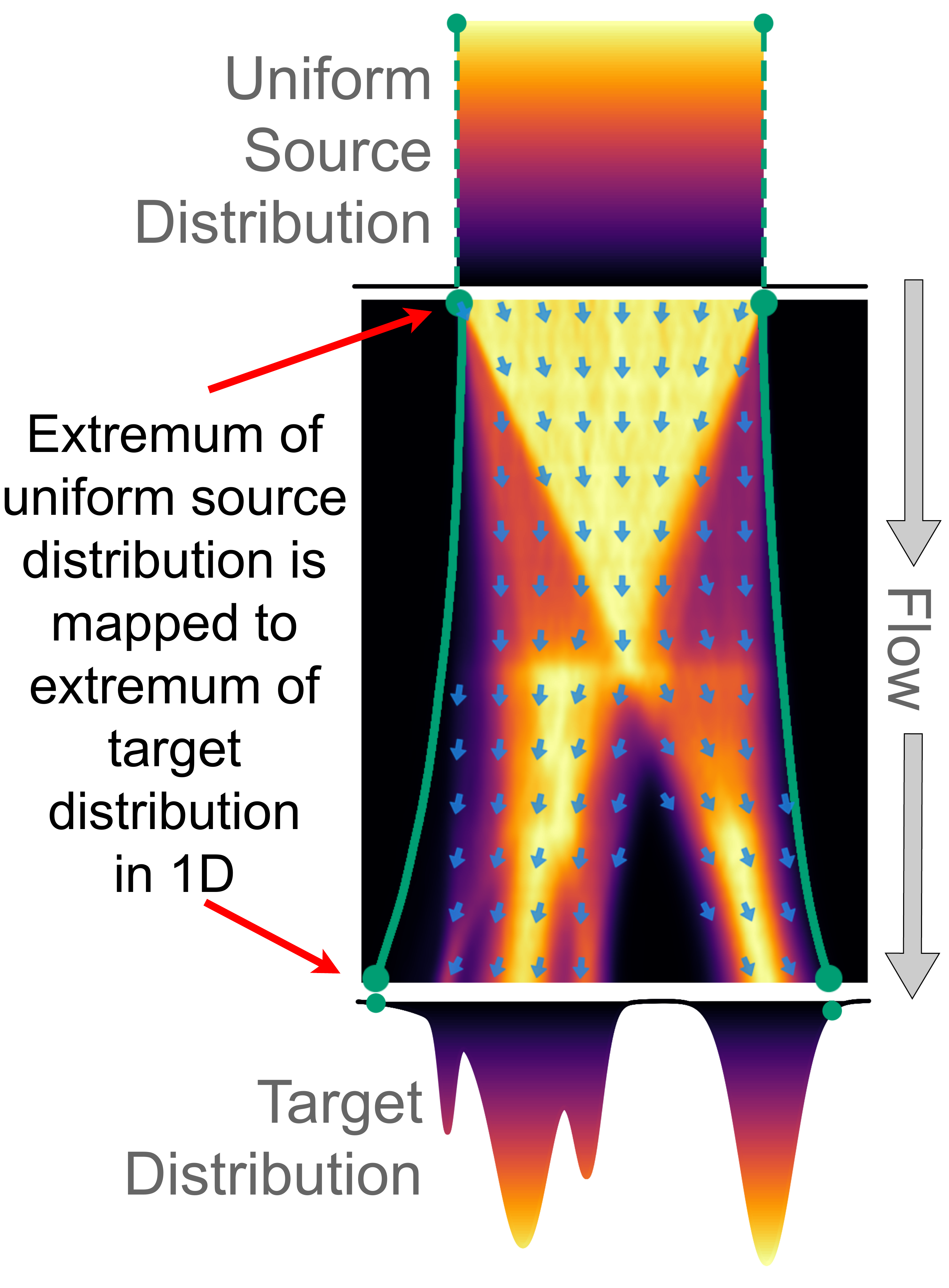}%
    \hspace{0.15cm}%
    \includegraphics[trim=0cm 0cm 0cm 0cm,clip,height=4.8cm]{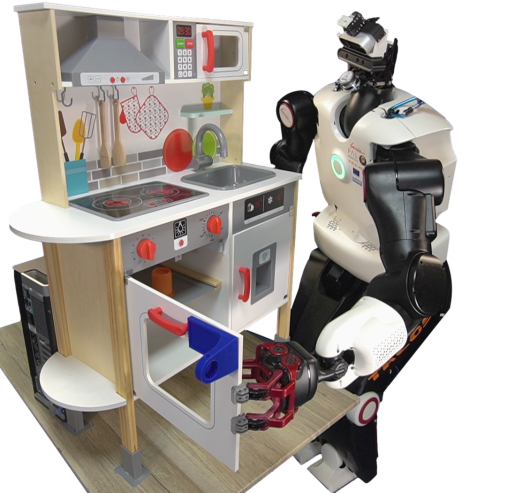}
    \caption{Goal conditioned policy using Extremum Flow Matching for manipulation from play demonstrations on the Talos humanoid robot (bottom right). The policy takes the current and goal images as input and outputs the robot's command trajectory (top). With Flow Matching, the flow vector field (bottom left) continuously and deterministically transforms a 1D source distribution into a target distribution. Since flow paths do not intersect, the boundaries of a uniform source distribution are mapped to the minimum and maximum of the target distribution's support. Using this idea, Extremum Flow Matching addresses the problem of optimality by selecting the shortest path to the goal.}
    \label{fig:concept}
\end{figure}

These recent advances in machine learning coincide with the rapid development of next-generation humanoid robot hardware \cite{cao2024ai, tong2024advancements}, driven by a similar long-standing ambition: to create generalist robotic platforms that combine versatile locomotion and dexterous manipulation skills, multi-contact capabilities, compact footprints suitable for constrained environments, and human-like morphology adapted to operate within human-centric spaces. The convergence of such hardware, traditional model-based control, and scalable, multimodal learning frameworks, acting as high-level controllers and planners, opens exciting new frontiers in humanoid robotics. 

In this work, we learn challenging long-horizon manipulation tasks in a realistic kitchen setting with a humanoid robot. As illustrated in Fig.~\ref{fig:concept} (bottom right) and accompanying videos, our learning algorithms can generate motions combining grasping, door opening/closing and tray pulling/pushing using the Talos humanoid robot, all based on high-dimensional visual input.

\begin{figure}[t]
    \centering
    \includegraphics[trim=0cm 0cm 0cm 0cm,clip,width=1.0\linewidth]{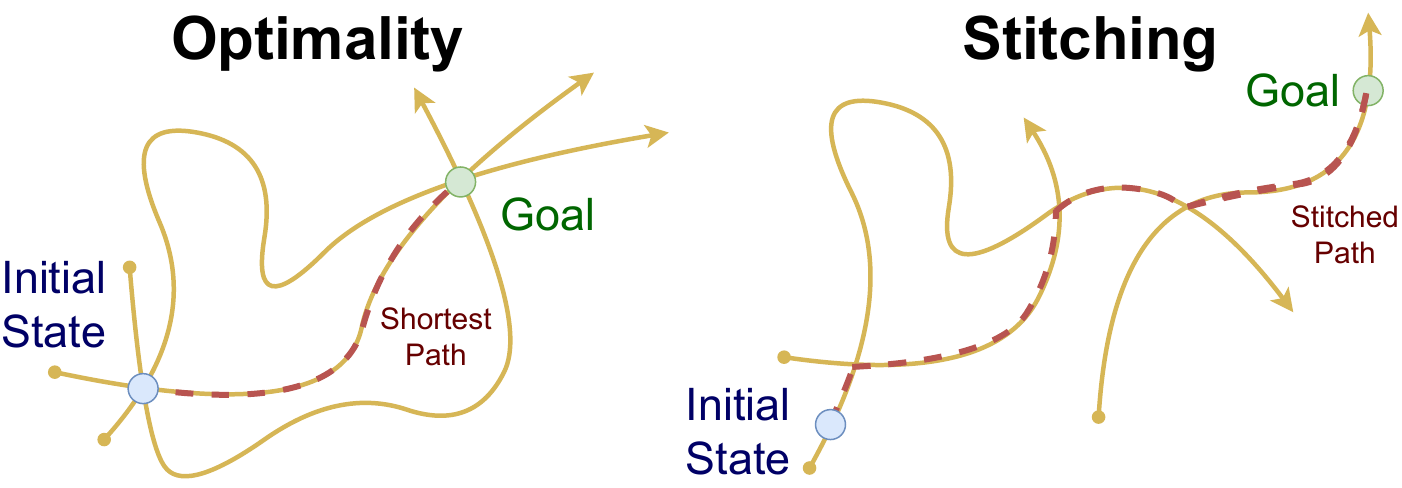}
    \caption{Imitation and offline reinforcement learning from play data face two key challenges: Optimality (left) -- selecting actions that lead to the shortest possible paths, and
    Stitching (right) -- combining sub-trajectories that originate from distinct demonstrated episodes to reach long-horizon tasks.}
    \label{fig:challenges}
\end{figure}

Imitation learning typically relies on near-optimal demonstrations focused on specific tasks to produce high-quality policies. However, scaling data collection under these constraints is costly and impractical. To overcome this, recent work has focused on learning from play datasets \cite{lynch2020learning, cui2022play, mees2022calvin, wang2023mimicplay}, in which demonstrations freely explore the environment through open-ended interactions without pursuing specific goals. This approach provides broader state-action coverage, greater variability, and makes data collection far more scalable and convenient, opening the door to large-scale, unstructured datasets collected from diverse sources. By conditioning on goals, it becomes possible to train from such play data, removing the need for carefully curated demonstrations. However, two major challenges remain (see Fig.~\ref{fig:challenges}): actions in play data are often suboptimal for specific tasks, and full trajectories from an initial state to a distant goal are rarely demonstrated. Although fragments of optimal behavior exist across episodes, they are not stitched into a complete solution. 

Optimality can be addressed with purely supervised learning techniques by conditioning on returns \cite{ajay2022conditional}. While such methods have shown interesting empirical performance, they lack a principled mechanism for stitching partial trajectories toward long-horizon goals. In contrast, offline reinforcement learning \cite{kostrikov2021offline,liu2024enhancing, park2023hiql, kim2024stitching, lu2025makes} explicitly tackles both optimality and stitching by learning value functions via dynamic programming.

These recent advances in imitation learning have been largely driven by progress in generative modeling, particularly Diffusion models \cite{ddpm, ddim} and Flow Matching \cite{flow}. Generative methods have proven decisive because they enable learning and sampling from the entire distribution, effectively handling multi-modal distributions. In contrast, traditional supervised learning often averages over non-convex distributions, leading to inaccurate predictions. Additionally, generative models can produce high-dimensional outputs, making them well-suited for generating full trajectories in planning. Addressing stitching capabilities, the application of Diffusion for long-term trajectory planning is an actively studied research direction \cite{janner2022planning, kim2024stitching, mishra2023generative, chen2024simple, luo2025generative, lu2025makes}.

Diffusion and flow-based methods have been unified within a common theoretical framework based on optimal transport theory \cite{albergo2023stochastic, albergo2023building}, where Diffusion corresponds to the stochastic formulation and Flow Matching to the deterministic one. Flow Matching has emerged as an alternative to Diffusion and has been applied to imitation and reinforcement learning \cite{rouxel2024flowmultisupport,hu2024adaflow,braun2024riemannian,zhang2025flowpolicy} as well as Vision-Language-Action models \cite{black2024pi_0}, primarily due to its faster inference times, a critical advantage for real-time robotic applications.

Specific to offline reinforcement learning, recent work exploited the unique properties of Flow Matching to address the key challenge of evaluating the critic value \textit{V} and action-value \textit{Q} functions. \cite{zhang2025energy} focused on improving the guidance of the generative process, while Flow Q-Learning \cite{park2025flow} introduced a distillation mechanism that integrates Flow Matching into reinforcement learning algorithms without requiring simulating the flow during training.

Concurrently with this line of work, we exploit the unique advantages of Flow Matching over Diffusion -- namely, its deterministic inference and its ability to handle arbitrary source distributions -- to address the reward maximization problem in learning the critic for reinforcement learning.

\textbf{Contributions:} Our key contributions are threefold:
\begin{itemize}
    \item To address the challenge of optimality and learn critic in offline reinforcement learning algorithms, we propose a method called \textit{Extremum Flow Matching}, which estimates distributional bounds using Flow Matching and conditioning on returns.
    \item We apply this method to design several goal conditioned imitation and offline reinforcement learning agents based on Flow Matching.
    \item We conduct extensive evaluations, comparing these agents against the state of the art using the OGBench benchmark, analyzing the impact of dataset collection behaviors, and validating performance through real-world experiments on the Talos humanoid robot (see attached video).
\end{itemize}

\section{Background}
\label{sec:preliminaries}

\subsection{Flow Matching}

Flow Matching is a generative method that enables to model and then sample from complex, potentially multi-modal probability distributions. It does so by learning a deterministic and continuous transformation from a simple source distribution $\bm{x}\up{src} \sim \mathcal{P}\up{src}$ to a target distribution $\bm{x}\up{dst} \sim \mathcal{P}\up{dst}$, where both $\bm{x}\up{src}$ and $\bm{x}\up{dst}$ lie in the same space $\R^n$. The flow's vector field $f$, typically parameterized by a neural network, is trained using supervised learning via the Flow Matching loss $\mathcal{L}\down{flow}$ without requiring simulation of the full integration process of the flow:
\begin{equation}\label{eq:flow_training}
\begin{aligned}
& \bm{x}_t = (1-t)\bm{x}\up{src} + t\bm{x}\up{dst},~ t \sim \mathcal{U}(0,1)\\
& \mathcal{L}\down{flow} = \norm{f(\bm{x}_t, t, \bm{c}) - (\bm{x}\up{dst} - \bm{x}\up{src})}_2^2
\end{aligned}
\end{equation}
where $t \in \R$ represents the interpolation, i.e. the progress of the transport from the source to the target distribution, and is sampled uniformly from the interval $[0,1]$. The vector $\bm{c}\in\R^m$ is an optional conditioning variable that allows modeling a conditional target distribution $P\up{dst}(\bm{x}|\bm{c})$.

During inference, novel samples from the target distribution $\tilde{\bm{x}}\up{dst}$ are generated by first sampling a point $\bm{x}_0$ from the source distribution $\mathcal{P}\up{src}$, and then integrating the learned flow from $t=0$ to $t=1$. This integration is typically performed using Euler integration:
\begin{equation}\label{eq:flow_inference}
\bm{x}_0 \sim \mathcal{P}\up{src},~ \bm{x}_{t+\Delta t} = \bm{x}_t + \Delta t f(\bm{x}_t,t,\bm{c}),~ \tilde{\bm{x}}\up{dst} = \bm{x}_1
\end{equation}

In the remainder of this paper, we denote the generative process $F$ conditioned by $\bm{c}$ as follows: the model definition and training process is expressed as $F: \mathcal{P}\up{src} | \bm{c} \mapsto \bm{x}\up{dst}$. Sampling-based inference is written as $\tilde{\bm{x}}\up{dst} = F(\mathcal{P}\up{src}|\bm{c})$, where a point is sampled from the source distribution $\mathcal{P}\up{src}$ and then transformed through the flow. Deterministic inference is denoted by $\tilde{\bm{x}}\up{dst} = F(\bm{x}_0|\bm{c})$, where given a specific source point $\bm{x}_0$, only the flow's integration from source to target is computed.

\subsection{Trajectory Dataset Formalism}
\label{sec:traj_notations}

In this work, we leverage the generative method Flow Matching to devise novel goal conditioned imitation and offline reinforcement learning algorithms. Rather than using the classical Markov Decision Process framework, where transitions are represented as $(\bm{o}_k,\bm{a}_k,\bm{o}_{k+1})$ and only consider immediate actions and subsequent observations, we adopt a trajectory-based formalism that captures sequences of observations and actions across episodes. 
Demonstrations are collected through teleoperation of the robot across multiple episodes, with each episode comprising a full trajectory of observations and corresponding actions. The set of demonstrated episodes is fixed and used in an offline reinforcement learning setting, where no further interaction with the environment occurs during training.

From the demonstrated episodes, we construct a trajectory-based training dataset composed of tuples of the form $(\bm{o}_k, \bm{\tau}^o_k, \bm{\tau}^a_k, d, \bm{g})$, where
$k\in\N$ denotes the time step within an episode, 
$\bm{o}_k$ is the observation at time step $k$,
$\bm{\tau}^o_k = \bm{o}_{k:k+L_oS_o:S_o}$ is the sub-sampled trajectory of future observations starting from $k$ of length $L_o$,
$\bm{\tau}^a_k = \bm{a}_{k:k+L_aS_a:S_a}$ is the sub-sampled trajectory of future actions starting from time step $k$ of length $L_a$,
$L_o$ and $L_a\in\N$ specify respectively the lengths of the future observation and future action trajectories,
$S_o$ and $S_a\in\N$ specify respectively the sub-sampling strides of future observation and action trajectories.
Akin to Hindsight Experience Replay \cite{andrychowicz2017hindsight}, the goal observation $\bm{g} = \bm{o}_{k+d}$ is sampled from the same episode, $d$ time steps after the current step $k$. The distance offset $d\in\N$ is drawn uniformly as $d\sim\mathcal{U}(0, L_g)$ where $L_g \in\N$ is the maximum goal horizon. Note that, when there is no ambiguity, the time step $k$ is omitted for clarity.

In this formalism, the return of an episode starting from observation $\bm{o}_k$ and reaching the goal $\bm{g}$ is defined as the time-step distance $d$, which the agent aims to minimize. In this work, we also assume that observations and goals lie in the same space, although prior work \cite{cui2022play, sridhar2024nomad,reuss2024multimodal,black2024pi_0} has demonstrated that goals can be encoded in alternative latent spaces or modalities, such as natural language.


\section{Method}
\label{sec:method}

\subsection{Problem Formulation}

Given a dataset of unstructured play demonstrations $\{(\bm{o}, \bm{\tau}^o, \bm{\tau}^a, d, \bm{g})_i\}$, our objective is to learn a policy $\pi: \bm{o},\bm{g}~\mapsto~\bm{\tau}^a$ that maps the current observation $\bm{o}$ and goal observation $\bm{g}$ to the next actions $\bm{\tau}^a$, such that the agent progresses toward the goal. However, achieving optimality (see Fig.~\ref{fig:challenges}) is challenging because the dataset contains both efficient and inefficient trajectories: some action sequences lead directly to the goal, while others take long detours, as the operator was not intentionally trying to reach a specific goal but merely \textit{passed by} it during demonstrations. The distribution $\mathcal{P}(\bm{\tau}^a,d\mid\bm{o},\bm{g})$ thus includes both optimal actions (associated with low $d$ values, representing more direct paths) and suboptimal ones (associated with higher $d$ values). The minimal distance between $\bm{o}$ and $\bm{g}$ within the dataset is given by $\min~\mathcal{P}(d \mid \bm{o}, \bm{g})$. To follow such shortest paths, the policy should condition on this minimal distance and select actions from the corresponding conditional distribution: $\mathcal{P}(\bm{\tau}^a|\bm{o}, \bm{g}, d=\min~\mathcal{P}(d|\bm{o},\bm{g}))$. In this sense, the problem of optimality reduces to learning and approximating the minimum (or maximum, in a reward-based formulation) of a conditional distribution.

Recent offline reinforcement learning methods \cite{kostrikov2021offline,liu2024enhancing, park2023hiql} have addressed this challenge of minimizing or maximizing conditional distributions using Expectile Regression \cite{newey1987asymmetric,kostrikov2021offline}. In this work, we propose an alternative approach based on Flow Matching, which enables the estimation of the lower and upper bounds of a conditional or unconditional distribution learned from offline data. Our method, along with recent work such as \cite{kim2024stitching, lu2025makes}, is inspired by the integration of reinforcement learning principles with conditioning on returns framework.

\subsection{Extremum Flow Matching}
\label{sec:flow_optim}

\begin{figure}[t]
    \centering
    \includegraphics[trim=0cm 0cm 0cm 0cm,clip,width=1.0\linewidth]{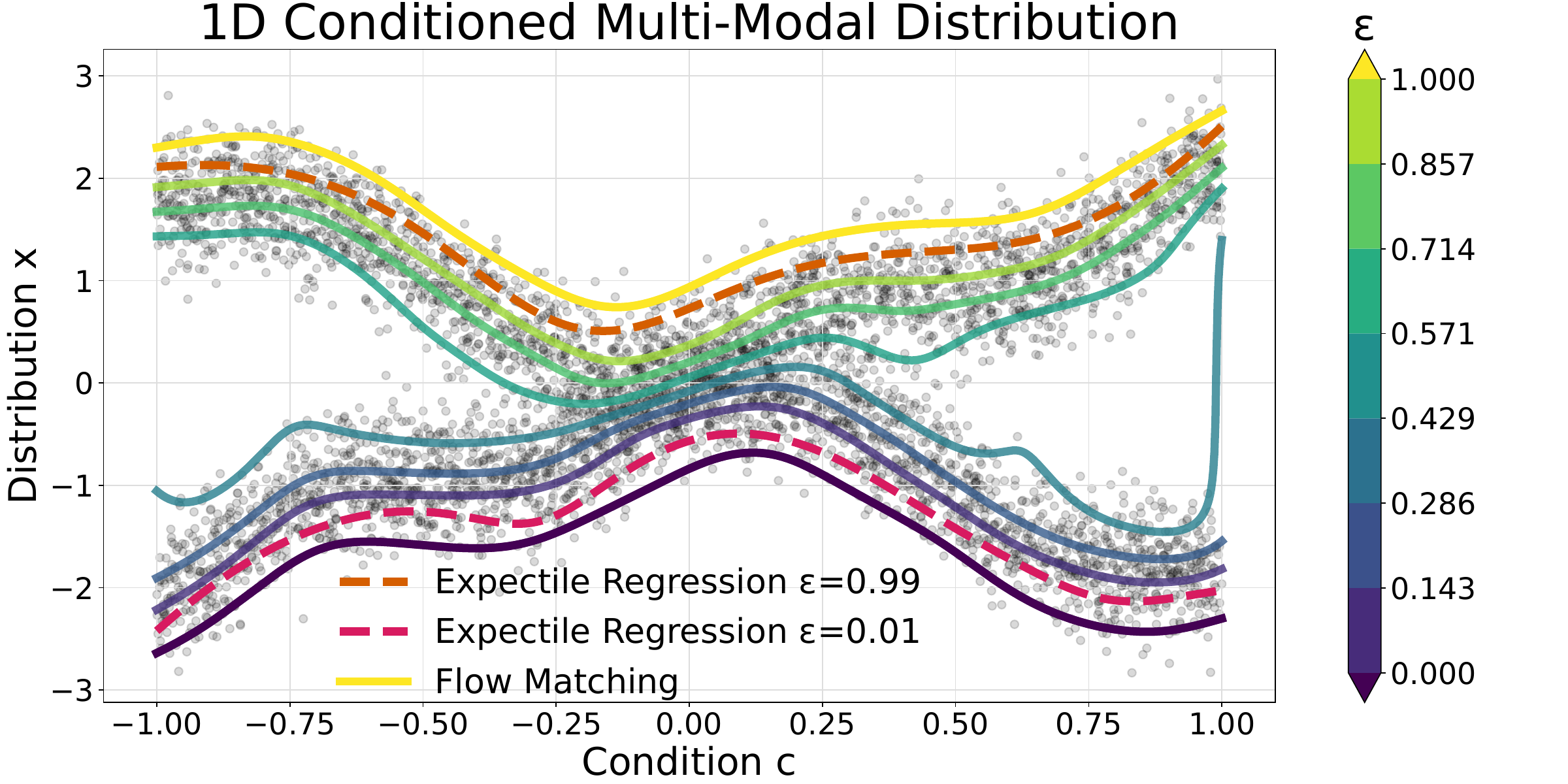}	
    \caption{Comparison between Expectile Regression and Extremum Flow Matching for estimating the minimum and maximum of a one-dimensional multi-modal conditional distribution. Extremum Flow Matching allows selecting the parameter $\epsilon$ at inference and provides tighter estimates.}
    \label{fig:flow_optim_1d}
\end{figure}

\textbf{One-Dimensional Distributions}: Fig.~\ref{fig:concept} (bottom left) illustrates the core concept of our approach. Flow Matching learns a transformation that is both continuous and deterministic, governed by a vector field whose integrated paths do not intersect. As a result, in the one-dimensional case, the minimum and maximum of the source distribution's support are mapped to the minimum and maximum of the target distribution.
Flow Matching offers key advantages over Diffusion-based methods: the transport is deterministic and allows the use of arbitrary source distributions, not just Gaussian ones. For this reason, we opt for a uniform distribution as the source which is not only easy to sample from but also provides a closed support with well-defined minimum and maximum bounds.

More formally, using the notation introduced earlier, a one-dimensional $x(\bm{c})\in\R$ distribution conditioned on $\bm{c}\in\R^m$ is learned by training the Flow Matching model $F:\epsilon\sim\mathcal{U}(0,1)|\bm{c} \mapsto x(\bm{c})$, where $\mathcal{U}(0,1)$ is the one-dimensional uniform distribution between $0$ and $1$. The minimum and maximum bounds are then inferred as $F(\epsilon=0|\bm{c})$ and $F(\epsilon=1|\bm{c})$, respectively. Note that if using a Gaussian source distribution, samples from the lower and upper tails can also be utilized to approximate the distribution's support bounds.

Fig.~\ref{fig:flow_optim_1d} shows a comparison between our Extremum Flow Matching approach and Expectile Regression on a one-dimensional conditional and multi-modal distribution. Expectile Regression estimates the bounds of the distribution by training the model $g$ using an asymmetric supervised loss, defined as $\mathcal{L}^\epsilon\down{expectile} = L_2^\epsilon(x - g(\bm{c}))$ where $L_2^\epsilon(u) = |\epsilon-\mathds{1}(u < 0)|u^2$. Expectile Regression approximates the minimum and maximum of the target distribution, respectively by setting $\epsilon=0.01$ and $\epsilon=0.99$.

While both methods perform well, Extremum Flow Matching, in contrast to Expectile Regression, learns to model the entire distribution, allowing the choice of $\epsilon$ at inference time rather than fixing it during training. We observed that Flow Matching tends to provide tighter estimates of the distributional bounds. Expectile Regression, on the other hand, generally yield more conservative approximations of the minimum and maximum, especially on multi-modal distributions.

\begin{figure}[t]
    \centering
    \includegraphics[trim=0cm 0cm 0cm 0cm,clip,width=1.0\linewidth]{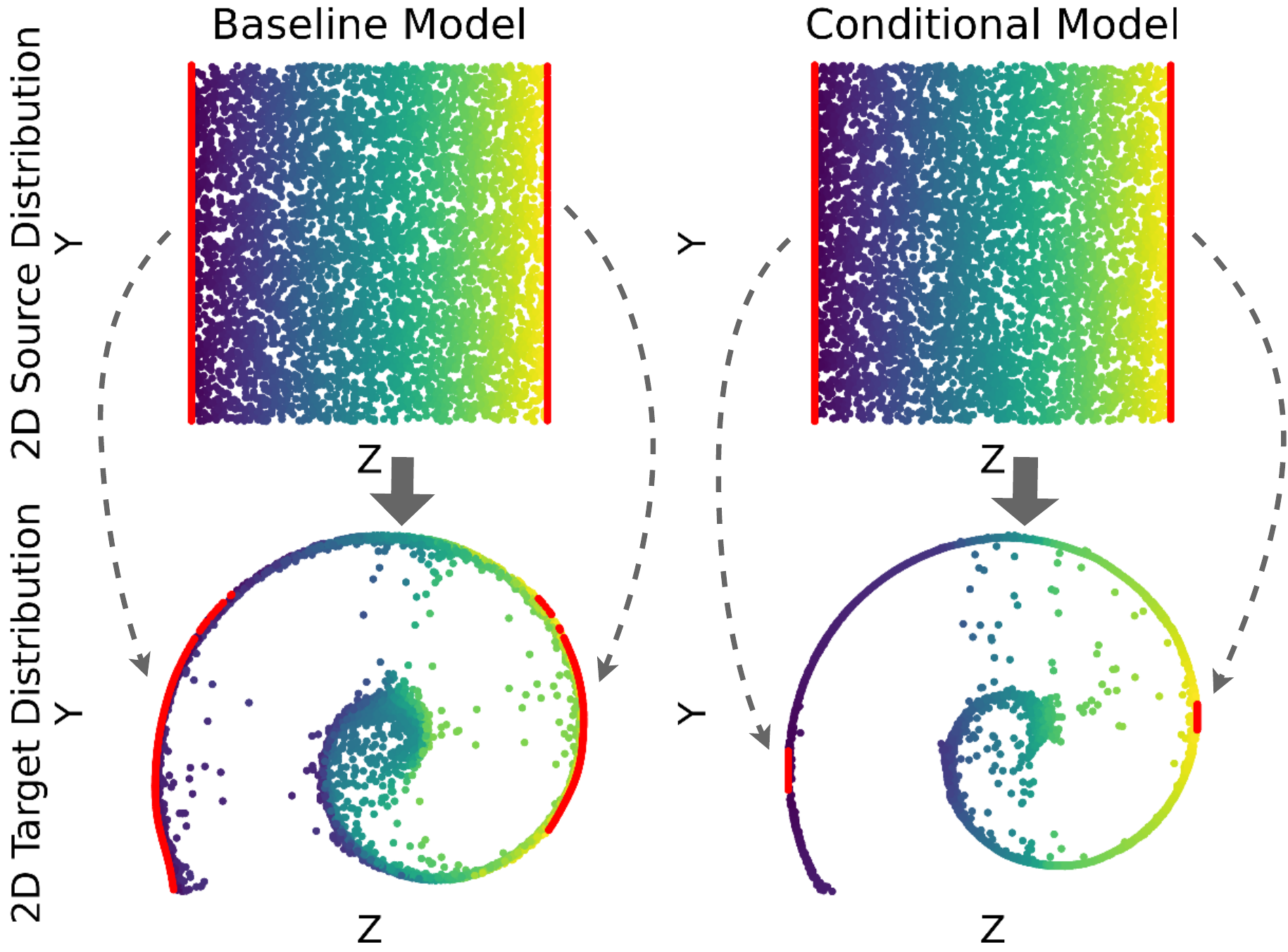}
    \caption{Extrema of unconditioned 2D distributions. Red samples illustrate how extremal values along the $z$ axis are mapped from the uniform source distribution (top) to the target distribution (bottom). The baseline approach using a single Flow Matching model (left) is compared to the proposed conditional model (right) described in Equations (\ref{eq:multidim_flow_train}) and (\ref{eq:multidim_flow_eval}). The baseline model fails to map the red extremal source samples to the true minimum and maximum along the $z$ axis of the target distribution, whereas the proposed conditional model achieves correct mappings.}
    \label{fig:flow_optim_2d}
\end{figure}

\textbf{Multi-Dimensional Distributions}: In the multi-dimensional setting where $\bm{x}\in\R^n$, we consider the problem of minimizing or maximizing the target distribution along a specific dimension $z\in\R$. We decompose the space as $\bm{x} = (z,\bm{y})$ with $z\in \R$ and $\bm{y} \in \R^{n-1}$. Formally, the objective is to sample from the conditional distribution $\mathcal{P}(z,\bm{y}|z=\text{max}~\mathcal{P}(z))$.

Since a distribution $\mathcal{P}$ can be decomposed as $\mathcal{P}(\bm{x}) = \mathcal{P}(z,\bm{y}) = \mathcal{P}(z)\mathcal{P}(\bm{y}|z)$, we propose to address the objective by training the following two models using Flow Matching:
\begin{equation}\label{eq:multidim_flow_train}
F_1: \mathcal{U}(0,1) \mapsto z,~~ F_2: \mathcal{P}\up{src}|z \mapsto \bm{y},
\end{equation}
where $\mathcal{P}\up{src}$ is an arbitrary source distribution that is easy to sample from for $\bm{y} \in \R^{n-1}$.

At inference time using these two generative processes, a sample $\tilde{\bm{x}}=(\tilde{z},\tilde{\bm{y}})$ corresponding to the minimization or maximization of the distribution along the $z$ axis is obtained with:
\begin{equation}\label{eq:multidim_flow_eval}
\tilde{z} = F_1(0) \text{ or } F_1(1),~~\tilde{\bm{y}} = F_2(\mathcal{P}\up{src}|\tilde{z}).
\end{equation}

This approach is illustrated in Fig.~\ref{fig:flow_optim_2d} (right). In contrast, the baseline with the single model $F(\mathcal{U}(0,1), \mathcal{U}(0,1)) \mapsto (z,y)$, shown on the left, demonstrates that without decomposing the generative process into two separate models, the extremal values along the $z$ axis in the source distribution are not reliably mapped to the corresponding extrema in the target distribution.

\subsection{Goal Conditioned Agents with Extremum Flow Matching}

We propose to use Extremum Flow Matching to devise a family of goal conditioned imitation learning and offline reinforcement learning algorithms. As pointed out in \cite{lu2025makes}, imitation learning and reinforcement learning agents are typically composed of multiple interacting components, which can generally be grouped into four main model types:
$\text{Critic}:~\bullet~\mapsto~d$ estimates the expected return (e.g., distance in time-step to goal) given an observation, goal, and/or action,
$\text{Planner}:~\bullet~\mapsto~\bm{\tau}^o$ \cite{janner2022planning, mishra2023generative, chen2024simple, luo2025generative} generates a single sub-goal or a trajectory of future observations that aims to reach the goal,
$\text{Actor}:~\bullet~\mapsto~\bm{\tau}^a$ produces a single action or a trajectory of actions,
$\text{World}:~\bm{\tau}^a,\bullet~\mapsto~\bm{\tau}^o$ (world model) \cite{ding2024diffusion, sobal2025learning} predicts the environment's dynamics by generating future observations from a trajectory of actions and current observation. 

Different algorithms make use of some, but not necessarily all, of these components. As shown in recent works \cite{lu2025makes,park2023hiql,sobal2025learning}, there are numerous possible combinations of these modules. However, the best performing configuration appears to be highly task and dataset dependent, and understanding the effect of these dependencies is still an open question. As noted by \cite{park2023hiql}, a key advantage of decomposing the agent into modular components is that certain models, such as the Planner, can be trained without action labels, potentially enabling the use of large-scale action-free datasets.

\begin{table*}[t]
\renewcommand{\arraystretch}{1.5}
\centering
\caption{Proposed Flow Matching Agents}
\label{table:agents}
\centering
\begin{tabular}{|l|p{3.9cm}|p{5.2cm}|p{6cm}|}
    \hline
    \textbf{Name} & \textbf{Training} & \textbf{Inference} & \textbf{Comment}
    \\ \hline
    \textbf{FM-GC} &
    $\text{Actor}: \mathcal{P}\up{src}_{\tau_a}|\bm{o},\bm{g} \mapsto \bm{\tau}^a$
    &
    $\tilde{\bm{\tau}}^a = \text{Actor}(\mathcal{P}\up{src}_{\tau_a}|\bm{o},\bm{g})$
    &
    \small Baseline goal conditioned with Flow Matching
    \\ \hline
    \textbf{FM-AC} &
    $\text{Critic}: \mathcal{U}(0,1)|\bm{o},\bm{g} \mapsto d$\newline
    $\text{Actor}: \mathcal{P}\up{src}_{\tau_a}|\bm{o},\bm{g},d \mapsto \bm{\tau}^a$
    &
    $\tilde{d} = \text{Critic}(0|\bm{o},\bm{g})$\newline
    $\tilde{\bm{\tau}}^a = \text{Actor}(\mathcal{P}\up{src}_{\tau_a}|\bm{o},\bm{g},\tilde{d})$
    &
    \small Actor conditioned, inspired by GCIQL \cite{kostrikov2021offline, park2023hiql} 
    \\ \hline
    \textbf{FM-PC} &
    $\text{Critic}: \mathcal{U}(0,1)|\bm{o},\bm{g} \mapsto d$\newline
    $\text{Planner}: \mathcal{P}\up{src}_{\tau_o}|\bm{o},\bm{g},d \mapsto \bm{\tau}^o$\newline
    $\text{Actor}: \mathcal{P}\up{src}_{\tau_a}|\bm{o},\bm{\tau}^o \mapsto \bm{\tau}^a$
    &
    $\tilde{d} = \text{Critic}(0|\bm{o},\bm{g})$\newline
    $\tilde{\bm{\tau}}^o = \text{Planner}(\mathcal{P}\up{src}_{\tau_o}|\bm{o},\bm{g},\tilde{d})$\newline
    $\tilde{\bm{\tau}}^a = \text{Actor}(\mathcal{P}\up{src}_{\tau_a}|\bm{o},\tilde{\bm{\tau}}^o)$
    &
    \small Planner conditioned, inspired by HIQL \cite{park2023hiql}
    \\ \hline
    \textbf{FM-PS} &
    $\text{Critic}: \mathcal{U}(0,1)|\bm{o},\bm{g} \mapsto d$\newline
    $\text{Planner}: \mathcal{P}\up{src}_{\tau_o}|\bm{o} \mapsto \bm{\tau}^o$\newline
    $\text{Actor}: \mathcal{P}\up{src}_{\tau_a}|\bm{o},\bm{\tau}^o \mapsto \bm{\tau}^a$
    &
    $T^o = \{\bm{\tau}^o | \bm{\tau}^o \sim \text{Planner}(\mathcal{P}\up{src}_{\tau_o}|\bm{o})\}$\newline
    $\tilde{\bm{\tau}}^o = \underset{\bm{\tau}^o \in T^o}{\argmin}~\text{Critic}(0|\bm{\tau}^o_{-1},\bm{g})$\newline
    $\tilde{\bm{\tau}}^a = \text{Actor}(\mathcal{P}\up{src}_{\tau_a}|\bm{o},\tilde{\bm{\tau}}^o)$
    &
    \small Planner rejection sampling, inspired by Diffusion Veteran \cite{lu2025makes}
    \\ \hline
    \textbf{FM-AS} &
    $\text{Critic}: \mathcal{U}(0,1)|\bm{o},\bm{g} \mapsto d$\newline
    $\text{Actor}: \mathcal{P}\up{src}_{\tau_a}|\bm{o} \mapsto \bm{\tau}^a$\newline
    $\text{World}: \mathcal{P}\up{src}_{\tau_o}|\bm{o},\bm{\tau}^a \mapsto \bm{\tau}^o$
    &
    $T^a = \{\bm{\tau}^a | \bm{\tau}^a \sim \text{Actor}(\mathcal{P}\up{src}_{\tau_a}|\bm{o})\}$\newline
    $\tilde{\bm{\tau}}^a = \underset{\bm{\tau}^a \in T^a}{\argmin}~\text{Critic}(0|\bm{\tau}^o_{-1},\bm{g})$\newline
    where $\bm{\tau}^o = \text{World}(\mathcal{P}\up{src}_{\tau_o}|\bm{o},\bm{\tau}^a)$
    &
    \small Actor rejection sampling with world model
    \\ \hline
\end{tabular}
\renewcommand{\arraystretch}{1.0}
\end{table*}

To further investigate and compare the influence of these different components, we introduce the set of agents described in Table~\ref{table:agents}. Here, $\mathcal{P}\up{src}_{\tau_a}$ and $\mathcal{P}\up{src}_{\tau_o}$ denote arbitrary source distributions for action and observation trajectories, and $\bm{\tau}^o_{-1}$ refers to the last observation of the trajectory $\bm{\tau}^o$. Note that by setting the trajectory lengths $L_a$ or $L_o$ to 1, this formalism includes and generalizes a wide range of previously proposed algorithms that either predict only the immediate next action or plan for a single sub-goal.

The agent \textbf{FM-GC} is a simple imitation learning policy baseline trained using Flow Matching without the Extremum idea of Section~\ref{sec:flow_optim}. While it is conditioned on the goal to generate action trajectories, it does not take the return $d$ into account. As a result, it lacks the notion of optimality and is theoretically incapable of stitching together partial trajectories across episodes to reach long-horizon goals.

Both \textbf{FM-AC} and \textbf{FM-PC} use the conditional scheme introduced in Section~\ref{sec:flow_optim} to estimate distribution extrema and address multi-dimensional distributions. They first train a Critic to model the distribution of returns $d$ given the current observation and goal. The optimal return (i.e., shortest time-step distance to goal) is inferred via $\text{Critic}(0|\bm{o},\bm{g})$ and used to condition the Actor in \textbf{FM-AC}, or the Planner in \textbf{FM-PC}. In \textbf{FM-PC}, the Actor plays the role of an inverse dynamics model, generating actions that realize the Planner’s predicted short-horizon observation trajectory.

These two agents represent Flow Matching counterparts to Return Conditioned Reinforcement Learning methods \cite{ajay2022conditional, kim2024stitching}, with a key distinction: prior approaches typically use hand-tuned, extreme return values (e.g., very low $d$ or high reward) as conditioning inputs, without estimating true optimal return based on observation and goal. This often pushes the model into out-of-distribution regions, relying on the network's ability to extrapolate. For instance, in 2D maze tasks, overly aggressive return values can lead to unrealistic plans, such as crossing walls. In contrast, our method estimates returns directly from the distribution seen in the training dataset, ensuring that the values used for conditioning are in-distribution.

Instead of conditioning on the returns, agents \textbf{FM-PS} and \textbf{FM-AS} follow the rejection sampling strategy benchmarked in Diffusion Veteran \cite{lu2025makes}. In \textbf{FM-PS}, the Planner, not conditioned on the goal, generates a set $T^o$ of candidate future observation trajectories from the current observation. At inference, the Critic evaluates the final observation $\bm{\tau}^o_{-1}$ of each candidate trajectory, and the one closest to the goal is selected. The Actor, trained as an inverse dynamics model, then produces the corresponding actions. Conversely in \textbf{FM-AS}, the Actor is used to samples a set $T^a$ of candidate action trajectories conditioned only on the current observation. These are passed through a learned World Model to predict resulting observations, and the Critic selects the best plan based on proximity to the goal.

\subsection{Reinforcement Learning Recursive Bootstrap}

While conditioning on returns or using rejection sampling can theoretically address action optimality, these methods are insufficient for enabling agents to stitch together trajectory segments across episodes. An ability that is essential for solving long-horizon tasks that are not fully demonstrated in the dataset. For each Critic-based agent in Table~\ref{table:agents}, we define two variants: \textbf{no-RL}, and \textbf{use-RL}. The \textbf{use-RL} variant augments the training batch using the following procedure: for each tuple $(\bm{o}, \bm{\tau}^o, \bm{\tau}^a, d, \bm{g})$ from the batch, an observation $\bm{g'}$ is uniformly sampled from the dataset, and the augmented tuple is added to the batch:
\begin{equation}\label{eq:rl}
\begin{aligned}
    & (\bm{o}, \bm{\tau}^o, \bm{\tau}^a, d+\text{Critic}(\epsilon_g|\bm{g},\bm{g'}), \bm{g'}), \\
    & \text{ with } \epsilon_g \sim \mathcal{U}(0,r_g), r_g\in[0,1],
\end{aligned}
\end{equation}
using the Critic model being trained. The term $d+\text{Critic}(\epsilon_g|\bm{g},\bm{g'}), \bm{g'})$ implements a Bellman-style backup by stitching together the returns of two trajectory segments: one from $\bm{o}$ to $\bm{g}$ with cost $d$, and another from $\bm{g}$ to $\bm{g'}$ with estimated return provided by the Critic. This effectively augments the return distribution with compositional trajectories, a structure that Flow Matching handles naturally. The additional goal $\bm{g'}$ can originate from a different episode \cite{kim2024stitching} than $\bm{o}$ and $\bm{g}$, such that the Critic is trained to estimate the return between any pair of observations in the dataset. The scaling factor $r_g \in [0,1]$ serves as a regularization hyperparameter to mitigate underestimation bias (overestimation of rewards) by introducing suboptimal return estimates into the distribution preventing collapse. The training and inference pseudocode for agent \textbf{FM-AC-use-RL} using high-dimensional image observations is detailed in Algorithm~\ref{algo:agent}. To further stabilize training, we also use the double networks trick \cite{hasselt2010double} on the Critic model.

\begin{algorithm}[t]
    \algsetup{linenosize=\footnotesize}
    \footnotesize
    \caption{FM-AC-use-RL with image observations}
    \label{algo:agent}
    \begin{algorithmic}[1]
        \STATE \textbf{Training:}
        \begin{ALC@g}
        \STATE \textbf{Input:} Training dataset
        \STATE Initialize Encoder, Critic and Actor models
        \WHILE{not converged}
            \FOR{$(\bm{o}, \bm{\tau}^a, d, \bm{g})$ sampled from batch}
                \STATE Sample observation $\bm{g'}$ from the whole dataset
                \STATE Encode observations into latent spaces:\newline$\bm{l}_o = \text{Encoder}(\bm{o})$, $\bm{l}_g = \text{Encoder}(\bm{g})$, $\bm{l}_{g'} = \text{Encoder}(\bm{g'})$
                \STATE Augment batch with $(\bm{l}_o, \bm{\tau}^a, d+\text{Critic}(\epsilon_g|\bm{l}_g,\bm{l}_{g'}), \bm{l}_{g'})$ (\ref{eq:rl})
            \ENDFOR
            \STATE Update Encoder and Critic model to minimize $\mathcal{L}\down{flow}$ (\ref{eq:flow_training})\newline with $x\up{src}=\mathcal{U}(0,1),x\up{dst}=d,c=(\bm{l}_o,\bm{l}_g)$
            \STATE Update Encoder and Actor model to minimize $\mathcal{L}\down{flow}$ (\ref{eq:flow_training})\newline with $x\up{src}=\mathcal{P}\up{src},x\up{dst}=\bm{\tau}^a,c=(\bm{l}_o,\bm{l}_g,d)$
        \ENDWHILE
        \STATE \textbf{Return:} Encoder, Critic and Actor models
        \end{ALC@g}
        \STATE \textbf{Inference:}
        \begin{ALC@g}
        \STATE \textbf{Input:} current $\bm{o}$ and goal $\bm{g}$ observations
        \STATE Encode observations into latent spaces:\newline$\bm{l}_o = \text{Encoder}(\bm{o})$, $\bm{l}_g = \text{Encoder}(\bm{g})$
        \STATE $\tilde{d} = \text{Critic}(0|\bm{l}_o,\bm{l}_g)$ using flow integration (\ref{eq:flow_inference})
        \STATE $\tilde{\bm{\tau}}^a = \text{Actor}(\mathcal{P}\up{src}_{\tau_a}|\bm{l}_o,\bm{l}_g,\tilde{d})$ using flow integration (\ref{eq:flow_inference})
        \STATE \textbf{Return:} action trajectory $\tilde{\bm{\tau}}^a$
        \end{ALC@g}
    \end{algorithmic}
\end{algorithm}


\section{Experimental Results}
\label{sec:result}

\begin{figure*}[t]
\footnotesize 
\setlength{\tabcolsep}{1.6pt}
\centering
    \begin{tabular}{|l|c|cccc|cccc|cccccc|}
        \hline
        \multicolumn{2}{|c|}{} & \multicolumn{4}{|c|}{\textbf{no-RL}} & \multicolumn{4}{|c|}{\textbf{use-RL}} & \multicolumn{6}{|c|}{\textbf{OGBench}}\\
        \hline
        \textbf{OGBench Dataset} & \textbf{FM-GC} & \textbf{FM-AC} & \textbf{FM-PC} & \textbf{FM-PS} & \textbf{FM-AS} & \textbf{FM-AC} & \textbf{FM-PC} & \textbf{FM-PS} & \textbf{FM-AS} & \textbf{GCBC} & \textbf{GCIVL} & \textbf{GCIQL} & \textbf{QRL} & \textbf{CRL} & \textbf{HIQL}\\
        \hline
        pointmaze-large-navigate-v0 & $66$ & $60$ & $60$ & $31$ & $29$ & $\bm{89}$ & $\bm{89}$ & $67$ & $64$ & $29$ & $45$ & $34$ & $86$ & $39$ & $58$\\
        pointmaze-large-stitch-v0 & $39$ & $37$ & $23$ & $15$ & $14$ & $40$ & $40$ & $42$ & $44$ & $7$ & $12$ & $31$ & $\bm{84}$ & $0$ & $13$\\
        antmaze-large-navigate-v0 & $7$ & $5$ & $5$ & $15$ & $1$ & $7$ & $22$ & $34$ & $15$ & $24$ & $16$ & $34$ & $75$ & $83$ & $\bm{91}$\\
        antmaze-large-stitch-v0 & $1$ & $0$ & $0$ & $6$ & $3$ & $0$ & $3$ & $18$ & $7$ & $3$ & $18$ & $7$ & $18$ & $11$ & $\bm{67}$\\
        cube-double-play-v0 & $\bm{69}$ & $32$ & $13$ & $22$ & $14$ & $2$ & $1$ & $12$ & $16$ & $1$ & $36$ & $40$ & $1$ & $10$ & $6$\\
        scene-play-v0 & $53$ & $52$ & $32$ & $42$ & $43$ & $7$ & $16$ & $40$ & $\bm{55}$ & $5$ & $42$ & $51$ & $5$ & $19$ & $38$\\
        puzzle-4x4-play-v0 & $1$ & $0$ & $3$ & $22$ & $\bm{48}$ & $0$ & $1$ & $14$ & $38$ & $0$ & $13$ & $26$ & $0$ & $0$ & $7$\\
        \hline
    \end{tabular}
    \caption{Comparison of our proposed agents on the OGBench benchmark \cite{ogbench_park2025}. We report the average of binary success rates (\%) across state-based observation datasets of intermediate difficulty. For each task, agents are evaluated on $5$ goals, and replicated $20$ times. We include for reference results from state-of-the-art algorithms reported in \cite{ogbench_park2025}.}
    \label{table:ogbench}
\end{figure*}

\subsection{Comparison on OGBench Benchmark}

We first evaluated and compared our proposed agents across a diverse suite of simulated locomotion and manipulation environments using low-dimensional, state-based observations from the OGBench benchmark \cite{ogbench_park2025}. OGBench is a recent project specifically targeting offline reinforcement learning, where training and evaluation datasets highlight multi-goal, stitching and combinatorial challenges. OGBench provides reference implementation and comparison of several state-of-the art algorithms.

Fig.~\ref{table:ogbench} presents the performance of our agents alongside several recent offline goal-conditioned reinforcement learning algorithms, comparing GCBC \cite{ghosh2019learning}, GCIVL and GCIQL \cite{kostrikov2021offline, park2023hiql}, QRL \cite{wang2023optimal}, CRL \cite{eysenbach2022contrastive}, and HIQL \cite{park2023hiql}. Unlike our approach, these baselines do not leverage generative methods nor the trajectory-based formalism. All of our agents were trained for $200$k epochs using the same set of hyperparameters across all tasks and agents. In contrast, the results reported from OGBench were obtained with hyperparameters being tuned for each task and baseline.

As observed in \cite{ogbench_park2025}, no single method consistently outperforms all others across every task. Our results reflect this observation: while some of our agents achieve strong performance on manipulation tasks, such as \textit{cube} and \textit{puzzle} requiring combinatorial stitching, they are comparatively less effective in certain locomotion tasks, especially \textit{antmaze} given fixed hypermarameters.

\subsection{Impact of Demonstration Behaviors}
\label{sec:compare_dataset}

\begin{figure}[t]
    \centering
    \includegraphics[trim=0cm 0cm 0cm 0cm,clip,width=0.47\linewidth]{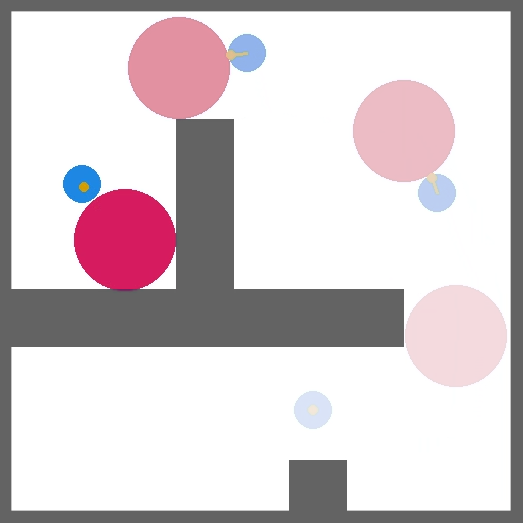}\hfill%
    \includegraphics[trim=0cm 0cm 0cm 0cm,clip,width=0.48\linewidth]{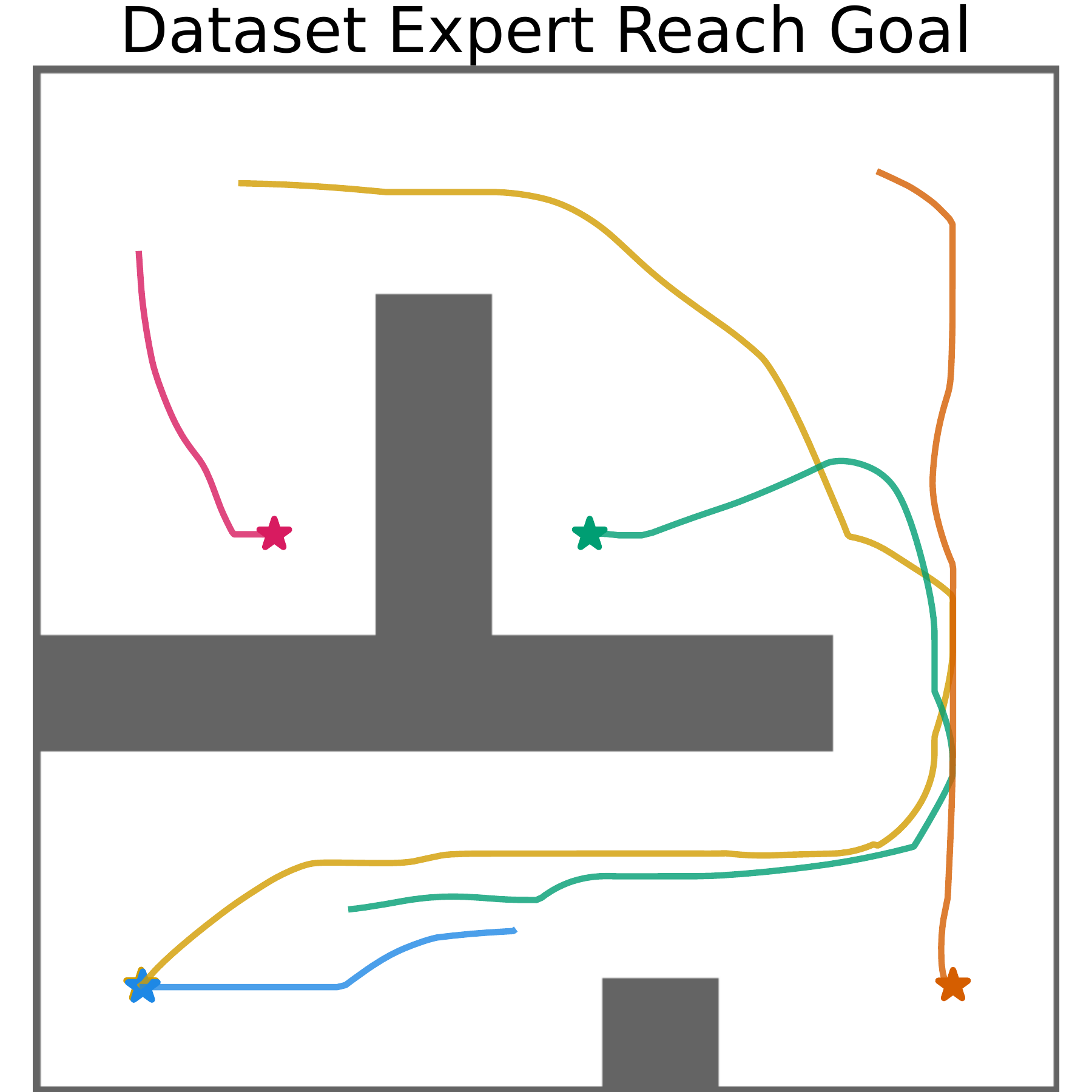}\\
    \vspace{0.3cm}%
    \includegraphics[trim=0cm 0cm 0cm 0cm,clip,width=0.48\linewidth]{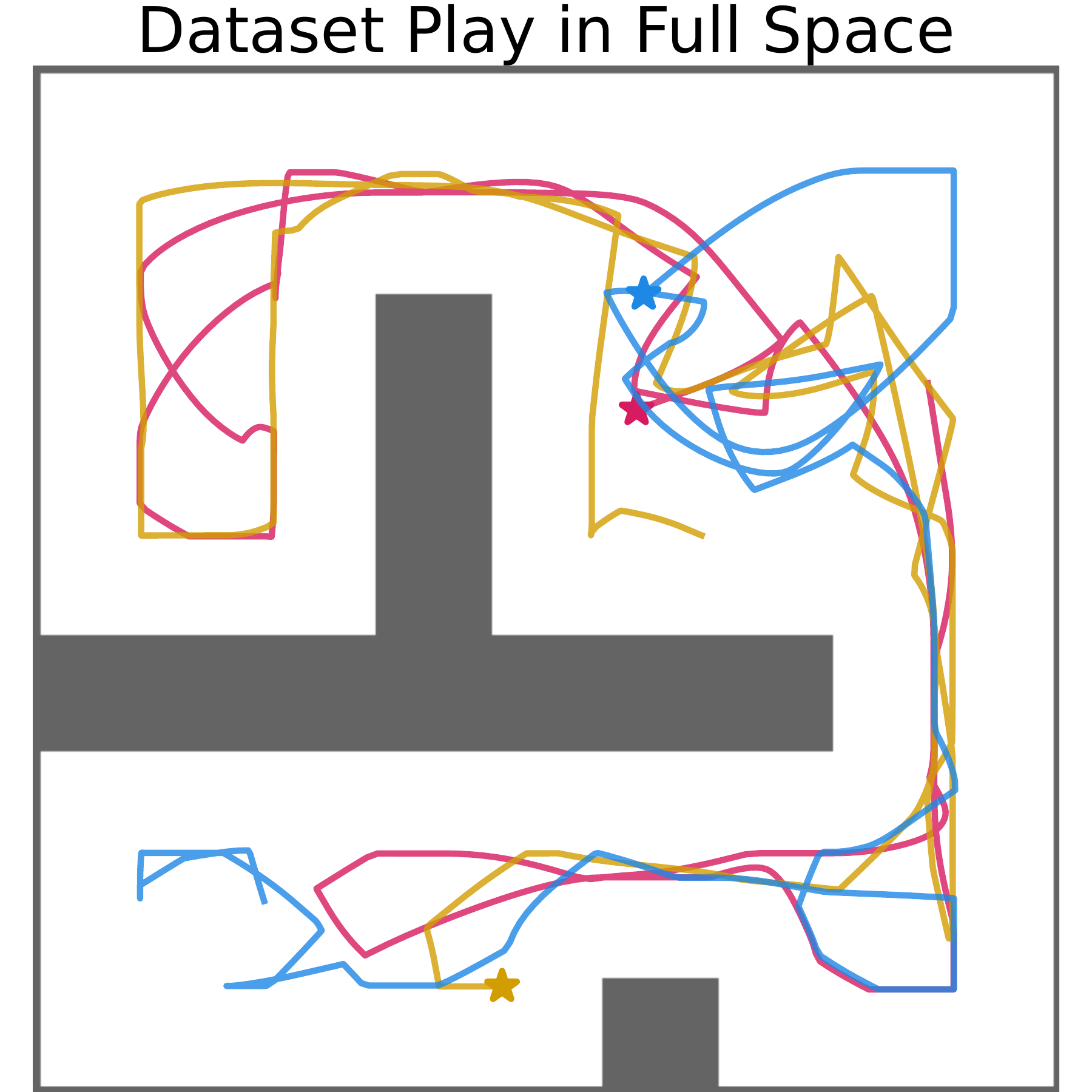}\hfill%
    \includegraphics[trim=0cm 0cm 0cm 0cm,clip,width=0.48\linewidth]{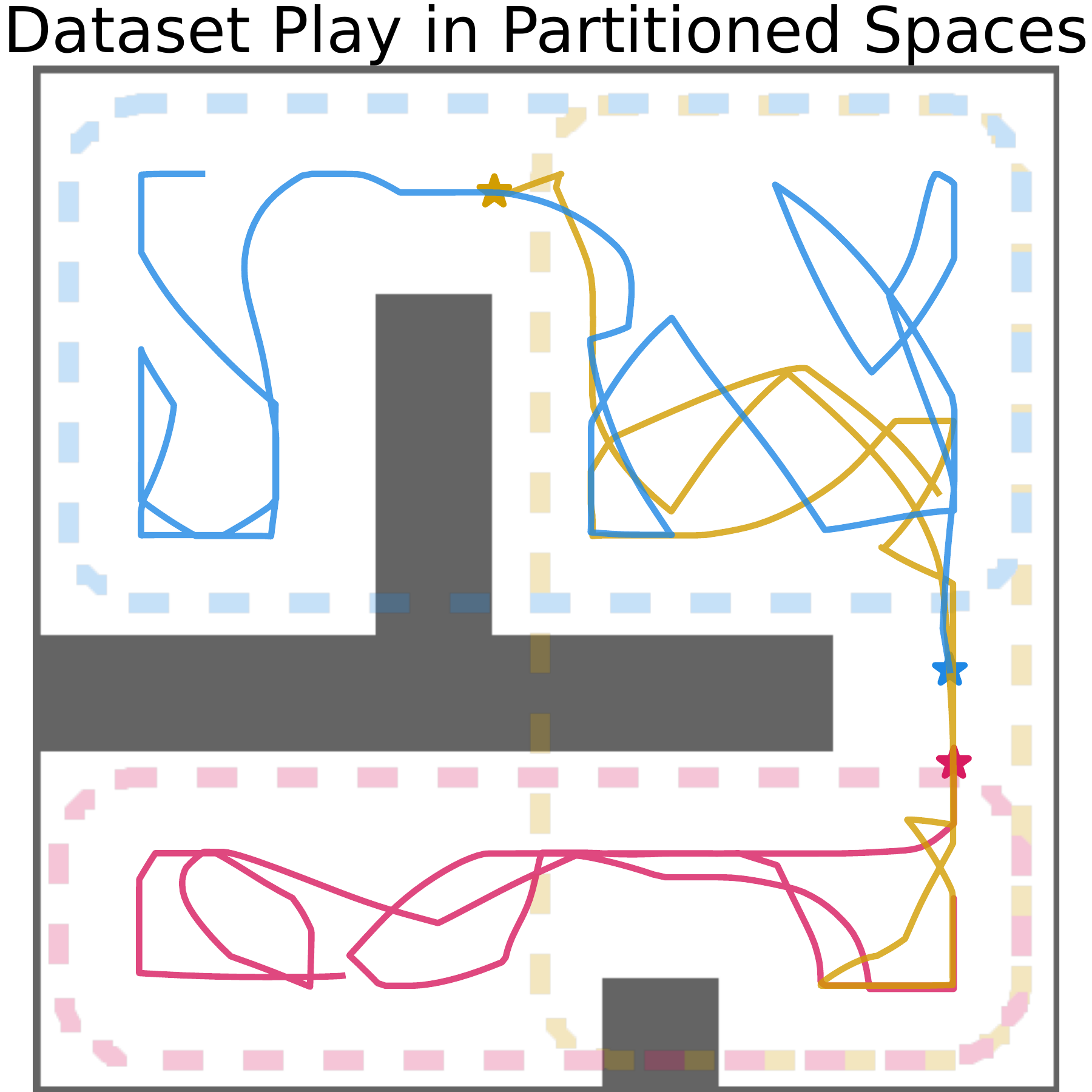}
    \caption{Planar pushing task in a maze (top left): the agent controls the small blue circle via position commands to move and push the larger passive red circle toward a corner of the maze, navigating around gray obstacle walls. Three datasets are recorded from human demonstrations exhibiting different behaviors. For each dataset, example episodes are shown, illustrating the motion trajectories of the red circle, with the star marker indicating the final position.}
    \label{fig:exp_planar}
\end{figure}

\begin{figure*}[t]
    \centering
    \begin{minipage}[c]{11.8cm}
        \includegraphics[trim=0cm 0cm 0cm 0cm,clip,width=\linewidth]{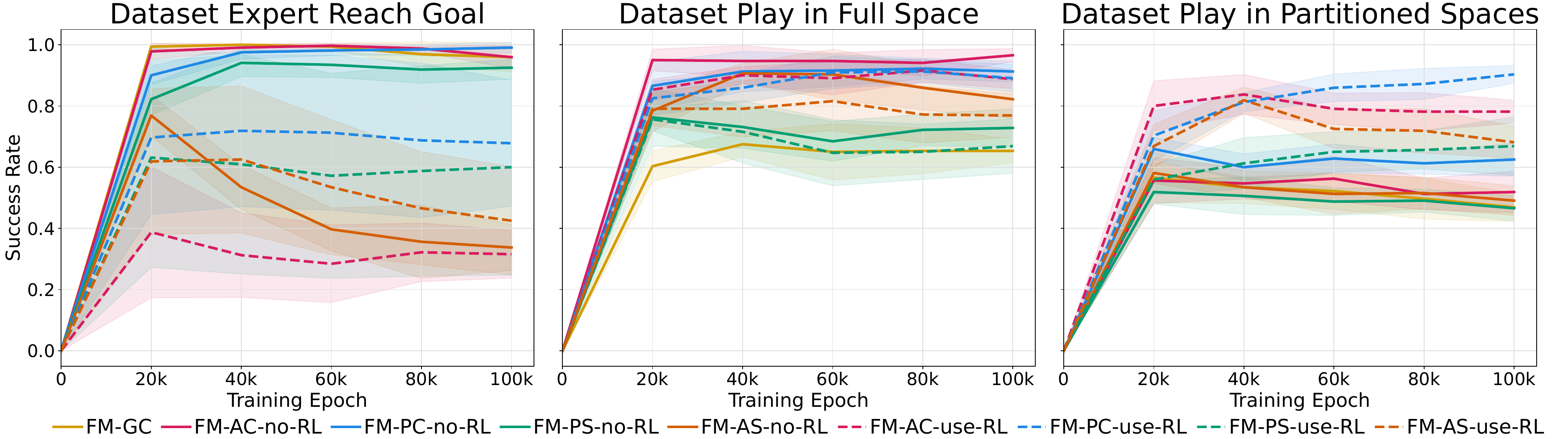}%
    \end{minipage}%
    \centering%
    \footnotesize%
    \setlength{\tabcolsep}{2pt}%
    \begin{tabular}{|l|ccc|}
        \hline
        \textbf{Agent} & \textbf{Expert} & \textbf{Full} & \textbf{Partitioned}\\
        \hline
        \textbf{FM-GC} & $\bm{96}${\scriptsize$\pm5$} & $65${\scriptsize$\pm4$} & $47${\scriptsize$\pm5$}\\
        \textbf{FM-AC-no-RL} & $\bm{96}${\scriptsize$\pm4$} & $\bm{97}${\scriptsize$\pm2$} & $52${\scriptsize$\pm7$}\\
        \textbf{FM-PC-no-RL} & $\bm{99}${\scriptsize$\pm2$} & $91${\scriptsize$\pm3$} & $62${\scriptsize$\pm5$}\\
        \textbf{FM-PS-no-RL} & $92${\scriptsize$\pm4$} & $73${\scriptsize$\pm6$} & $47${\scriptsize$\pm4$}\\
        \textbf{FM-AS-no-RL} & $34${\scriptsize$\pm8$} & $82${\scriptsize$\pm6$} & $49${\scriptsize$\pm5$}\\
        \textbf{FM-AC-use-RL} & $32${\scriptsize$\pm8$} & $89${\scriptsize$\pm3$} & $78${\scriptsize$\pm4$}\\
        \textbf{FM-PC-use-RL} & $68${\scriptsize$\pm21$} & $89${\scriptsize$\pm4$} & $\bm{90}${\scriptsize$\pm3$}\\
        \textbf{FM-PS-use-RL} & $60${\scriptsize$\pm35$} & $67${\scriptsize$\pm9$} & $67${\scriptsize$\pm10$}\\
        \textbf{FM-AS-use-RL} & $42${\scriptsize$\pm18$} & $77${\scriptsize$\pm7$} & $68${\scriptsize$\pm5$}\\
        \hline
    \end{tabular}
    \caption{Success rates of proposed agents across three datasets with varying demonstration behaviors. Performance is evaluated on the same set of $10$ initial-goal observation pairs, averaged over $8$ random training seeds and 4 runs per evaluation pair. The plots (left) show the evolution of success rates during training, while the table (right) presents the final success rates with standard deviation.}
    \label{fig:compare_dataset}
\end{figure*}

To further investigate these benchmark results, we studied the impact on agent's performance of demonstration behavior, i.e. the strategy, style, or consistency with which humans performed the task during data collection. We compared our agents on the planar pushing task illustrated in Fig.~\ref{fig:exp_planar} (top left), using three datasets collected from human demonstrations with distinct behaviors. All agents were trained with identical hyperparameters and evaluated on the same initial-goal observation pairs.

As shown in Fig.~\ref{fig:exp_planar}, each dataset reflect a different behavior:
In \textit{Expert Reach Goal} dataset ($46793$ samples), each episode demonstrates an optimal trajectory where the red circle is pushed from a random initial state to one of the maze corners, emphasizing goal directed expert behavior.
In \textit{Play in Full Space} dataset ($36661$ samples), longer, exploratory play episodes move freely the red circle throughout the maze without targeting specific goals, capturing diverse but suboptimal behaviors.
In \textit{Play in Partitioned Spaces} dataset ($29087$ samples), episodes also consist of exploratory play but are confined to one of the three maze regions, shown as dashed lines in Fig.~\ref{fig:exp_planar} (bottom right), with no single trajectory spanning all regions. This dataset highlights agents' trajectory stitching abilities, as reaching distant evaluation goals requires crossing all three regions, a path that was never demonstrated in the dataset.

Evaluation results are shown in Fig~\ref{fig:compare_dataset}. As expected, \textit{Expert Reach Goal} is the easiest dataset, followed by \textit{Play in Full Space} with sub-optimal demonstrations, and \textit{Play in Partitioned Spaces}, which is the most difficult due to the need for trajectory stitching. No single agent consistently outperforms others across all settings, aligning with previous findings. Notably, using RL backups (Eq.~\ref{eq:rl}) significantly degrades performance on the expert dataset, slightly reduces it on full-space play, but is crucial for solving partitioned-space play, where stitching is required. The underlying cause of this trade-off remains unclear and is a key direction for future research. On the expert dataset, simple goal-conditioned imitation with Flow Matching baseline performs comparably to more complex agents (without RL backups). However, in the more challenging play datasets, agents with Critics considering optimality outperform the baseline. The agent \textbf{FM-AS} using a world model performed poorly on the expert dataset regardless of whether RL backups were used. Its performance consistently degraded over the course of training, a phenomenon we plan to investigate further.

\begin{figure}[t]
    \centering
    \includegraphics[trim=4.5cm 0cm 12cm 0cm,clip,height=5cm]{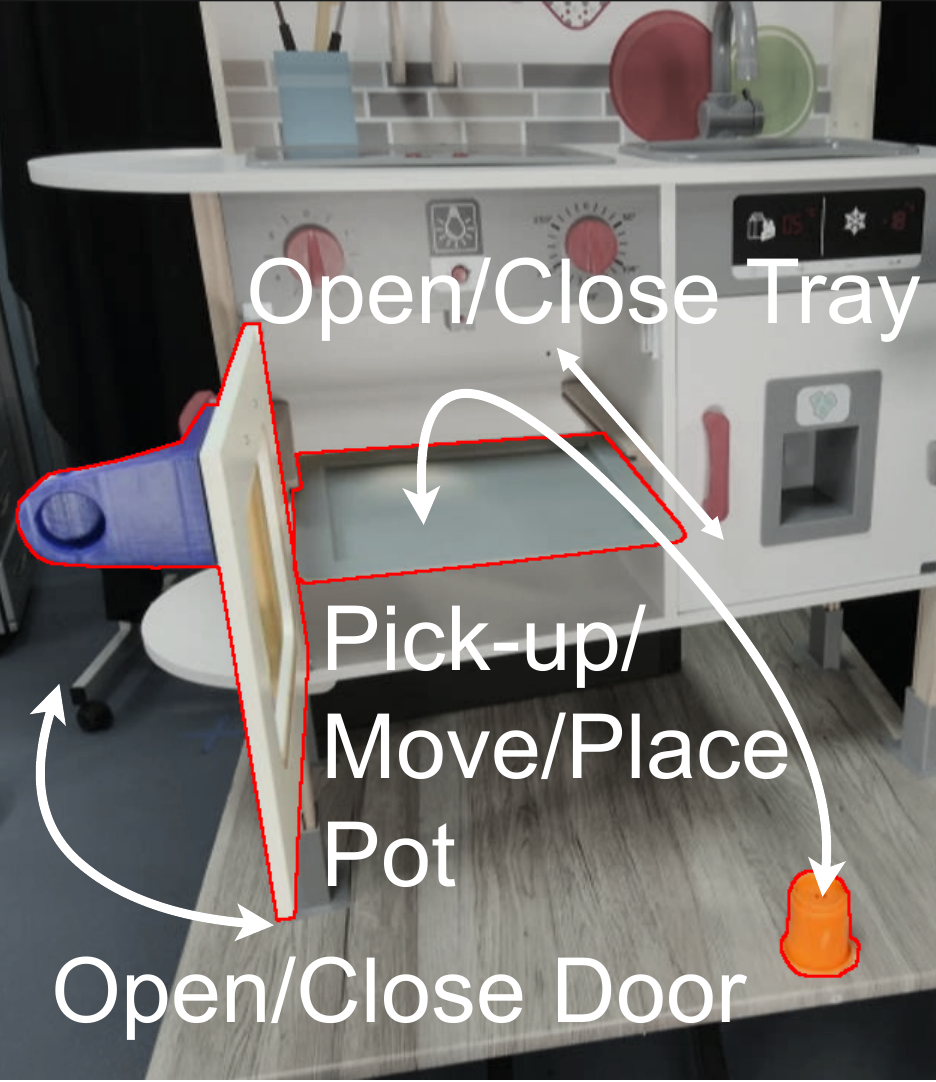}
    \includegraphics[trim=0cm 0cm 0cm 0cm,clip,height=5cm]{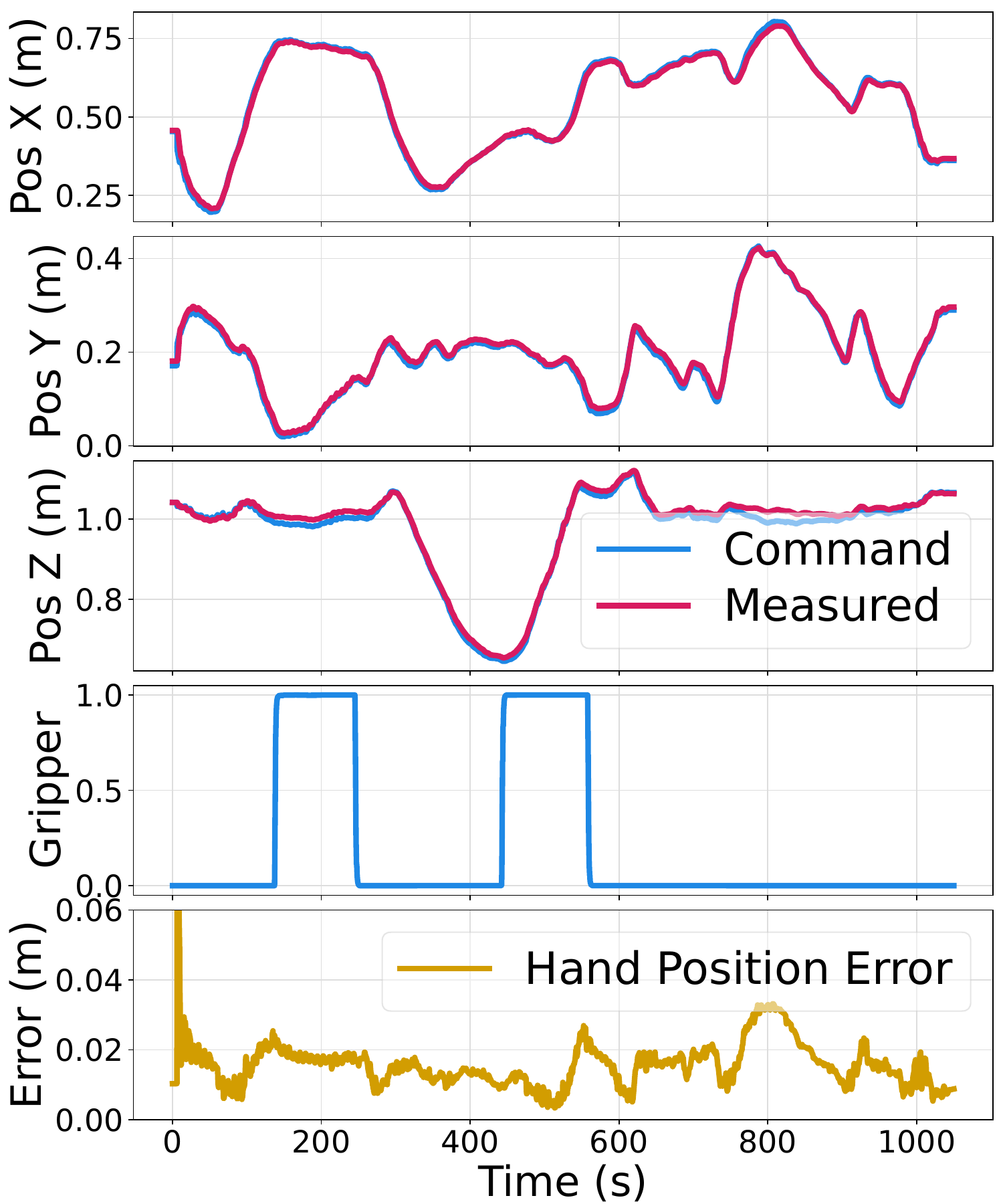}
    \caption{Kitchen task setup for image-based experiments on the Talos humanoid robot (left), and example of autonomous policy execution (right): left-hand position (commanded and measured), gripper command (open=0, close=1) and hand position error in world frame are displayed. The robot grasps and pulls open the tray, lowers its hand to grasp the pot on the table, places it on the tray and closes the tray by pushing it (see attached video).}
    \label{fig:exp_talos_detail}
\end{figure}

\subsection{Vision-Based Manipulation with Talos Humanoid Robot}

\begin{figure*}[t]
    \centering
    \includegraphics[trim=0cm 0cm 0cm 0cm,clip,width=0.24\linewidth]{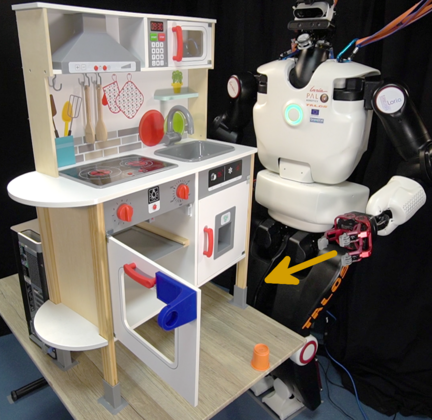}%
    \hspace{0.1cm}%
    \includegraphics[trim=0cm 0cm 0cm 0cm,clip,width=0.24\linewidth]{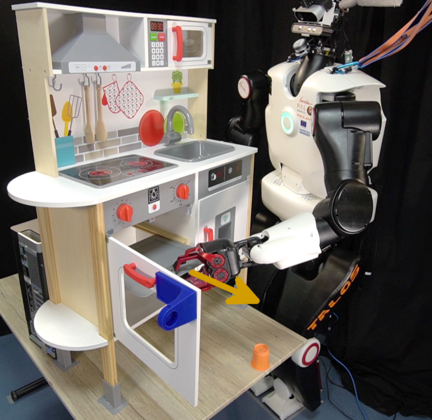}%
    \hspace{0.1cm}%
    \includegraphics[trim=0cm 0cm 0cm 0cm,clip,width=0.24\linewidth]{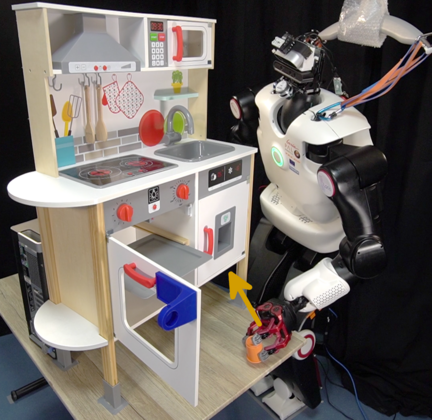}%
    \hspace{0.1cm}%
    \includegraphics[trim=0cm 0cm 0cm 0cm,clip,width=0.24\linewidth]{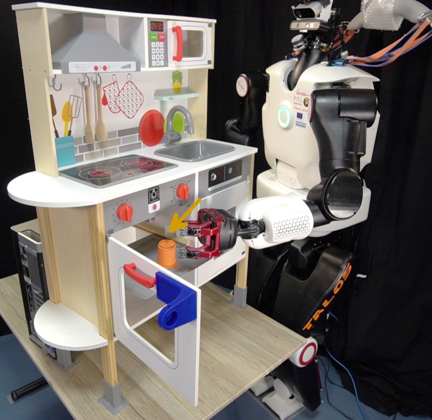}\\
    \vspace{0.1cm}
    \includegraphics[trim=0cm 0cm 0cm 0cm,clip,width=0.24\linewidth]{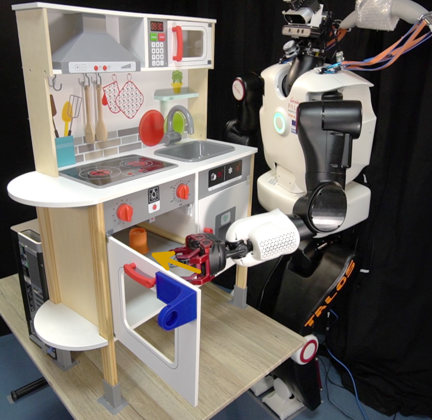}%
    \hspace{0.1cm}%
    \includegraphics[trim=0cm 0cm 0cm 0cm,clip,width=0.24\linewidth]{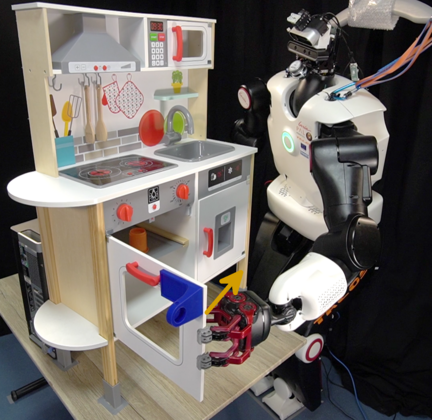}%
    \hspace{0.1cm}%
    \includegraphics[trim=0cm 0cm 0cm 0cm,clip,width=0.24\linewidth]{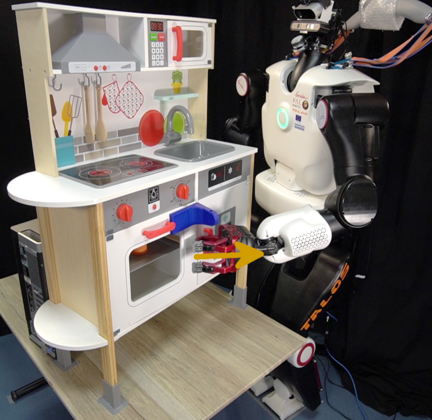}%
    \hspace{0.1cm}%
    \includegraphics[trim=0cm 0cm 0cm 0cm,clip,width=0.24\linewidth]{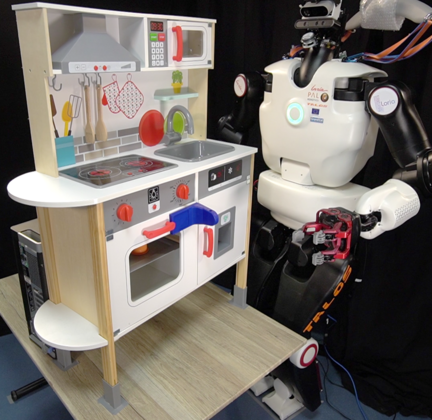}
    \caption{When conditioned on the current and desired goal images shown in Fig.~\ref{fig:concept} (top), the \textbf{FM-AC-no-RL} policy commands the Talos humanoid robot to sequentially open the tray, pick up the pot from the table, place it on the tray, close the tray, and shut the door to reach the goal.
    }
    \label{fig:exp_talos_video}
\end{figure*}

We further validated our method with high-dimensional image observations on real hardware using the Talos humanoid robot, featuring 24 degrees of freedom plus a gripper-equipped hand. Talos is controlled via SEIKO whole-body admittance controller\footnote{SEIKO implementation: \url{https://github.com/hucebot/seiko_controller_code}} based on Quadratic Programming optimization \cite{rouxel2024multicontact,seiko,seiko2}, which receives Cartesian hand pose commands in the robot's world frame and computes joint position references while maintaining balance constraints.

The robot interacts with a realistic kitchen environment as shown in Fig~\ref{fig:concept} (bottom right), and Fig.~\ref{fig:exp_talos_detail} (left). Using one hand, the robot picks up and places a red pot, opens and closes an articulated door, and operates a pull-out tray. The pot is placed either on the table or on the tray. Observations include RGB images from a fixed external camera and proprioceptive data: measured and commanded hand positions and orientations. The pot remains visible from the camera even when inside the kitchen's cabinet and when the door is closed. Images are encoded using a ResNet-based small IMPALA encoder \cite{espeholt2018impala} trained from scratch. Hand orientations are represented using the 6D continuous representation \cite{zhou2019continuity}. The goal conditioned policy takes the current and goal observations as input, defining the desired state of the environment, and outputs a trajectory of future actions, including hand position/orientation and gripper commands. Demonstrations were collected via teleoperation using a Vive pose tracking device, covering $32$ minutes over multiple episodes at $33$Hz. Action trajectories are sub-sampled at $11$Hz ($S_a=3$) with a length of $L_a=16$ steps, resulting in trajectories of $1.45$s long. The policy runs on an external machine with an NVIDIA GTX 1080 GPU, achieving about $0.14$ms inference time and recomputing trajectory updates every $0.8$s.

During demonstrations, the human operator performed unstructured play, manipulating the environment without a fixed task or sequence, opening/closing the door and tray, and moving the pot freely. After training, the policy is able to reach specific visual goal observations from an initial observation. One such rollout is shown in Fig.~\ref{fig:exp_talos_detail} (right) and Fig.~\ref{fig:exp_talos_video}: the robot opens the tray, places the pot inside the kitchen, closes the tray, and then closes the door to match the provided goal image. We compared qualitatively the three agents \textbf{FM-GC}, \textbf{FM-AC-no-RL}, and \textbf{FM-AC-use-RL}. Consistent with the results in Section~\ref{sec:compare_dataset}, \textbf{FM-AC-no-RL} demonstrated the best performance and was used to generate the accompanying videos\footnote{Project webpage: \url{https://hucebot.github.io/extremum_flow_matching_website/}}. Real-world success rate reaches 30\%, primarily due to hand motion inaccuracies during critical phases like door manipulation and object grasping, which require sub-centimeter precision. The whole-body controller introduces slight hand positioning errors (see Fig.~\ref{fig:exp_talos_detail} right bottom row) due to closed-loop admittance control based on force-torque sensor feedback. As the foot force-torque sensors tend to drift over time, these errors are not always repeatable between demonstrations and evaluations. Although these effects are measurable and included in the proprioceptive observation, the policy does not generalize well and adapt given the limited demonstration data.


\section{Limitations}
\label{sec:discussion}

Compared to algorithms such as GCBC, GCIVL, GCIQL, QRL, CRL, and HIQL implemented in OGBench, which do not rely on generative models, our agents require significantly more training time. This overhead is further increased when using the RL backup variants, as the batch augmentation procedure (Eq.~\ref{eq:rl}) involves integrating the flow model to infer the Critic, adding substantial computational cost. However, the increased computation time is justified by Flow Matching’s ability to model multi-modal distributions of actions, plans, and sub-goal \cite{rouxel2024flowmultisupport}, crucial for scaling to more complex tasks and larger datasets .

In our comparison, we evaluated all proposed agents using the same set of hyperparameters to enable a fair assessment of their robustness to parameter finetuning. However, we did not conduct a comprehensive exploration of hyperparameter sensitivity, which may influence the relative performance rankings on both the OGBench benchmark and the planar pushing task.

Regarding real-world deployment on the Talos humanoid, the policy was not able to generalize well to out of distribution cases, for instance, grasping the pot when it was positioned differently from the training demonstrations. This limitation is likely due in part to the small size of the training dataset, which constrained the policy’s ability to learn robust generalization.


\section{Conclusion}
\label{sec:conclusion}

We introduced a novel method, Extremum Flow Matching, designed to estimate the minimum and maximum of a multivariate probability distribution along a specific axis. This is made possible by the unique property of Flow Matching to utilize a uniform distribution as the source, combined with a conditioning scheme analogous to conditioning on returns. Building on this foundation, we proposed several imitation learning and offline reinforcement learning agents based on Flow Matching, varying in their internal architecture and module combinations.

Through simulation experiments, some of our agents outperformed state-of-the-art baselines on specific tasks within the OGBench benchmark, while underperforming on others, mirroring the task-specific variability seen in existing methods. Our findings reinforce that algorithm performance is highly sensitive to the data collection policy, as datasets can exhibit markedly different properties depending on the behavior used to generate them. Understanding and mitigating this sensitivity is crucial for developing more robust and generalizable agents capable of performing reliably across a diverse range of tasks.

We demonstrated the real-world applicability of our approach by deploying the proposed goal-conditioned agents on the full-size Talos humanoid robot. The learned vision-based policy interfaces with a low-level, model-based whole-body controller, enabling the robot to perform complex manipulations in a kitchen environment. Notably, the policy was trained using unstructured, teleoperated play demonstrations. This showcases the potential of leveraging large-scale, suboptimal datasets to train more generalist humanoid robots.


\bibliographystyle{IEEEtran}
\bibliography{references}

\end{document}